\documentclass[10pt]{cai}

\begin{document}
\def\conferenceyear{2025}
\volumeheader{38}{0}
\begin{center}

\title{Agile Reinforcement Learning for Real-Time Task Scheduling in Edge Computing}
\maketitle

\thispagestyle{empty}

\begin{tabular}{cc}
Amin Avan\upstairs{\affilone,*}, Akramul Azim\upstairs{\affilone}, Qusay H. Mahmoud\upstairs{\affilone}
\\[0.25ex]
{\small \makebox[\linewidth][c]{%
\parbox{0.9\linewidth}{\centering \upstairs{\affilone}Department of Electrical, Computer and Software Engineering, Ontario Tech University, Oshawa, Ontario, Canada}}}
\end{tabular}
  
\emails{
  \upstairs{*}amin.avan@ontariotechu.net 
}
\end{center}

\begin{abstract}
Soft real-time applications are becoming increasingly complex, posing significant challenges for scheduling offloaded tasks in edge computing environments while meeting task timing constraints. Moreover, the exponential growth of the search space, presence of multiple objectives and parameters, and highly dynamic nature of edge computing environments further exacerbate the complexity of task scheduling. As a result, schedulers based on heuristic and metaheuristic algorithms frequently encounter difficulties in generating optimal or near-optimal task schedules due to their constrained ability to adapt to the dynamic conditions and complex environmental characteristics of edge computing. Accordingly, reinforcement learning algorithms have been incorporated into schedulers to address the complexity and dynamic conditions inherent in task scheduling in edge computing. However, a significant limitation of reinforcement learning algorithms is the prolonged learning time required to adapt to new environments and to address medium- and large-scale problems. This challenge arises from the extensive global action space and frequent random exploration of irrelevant actions. Therefore, this study proposes Agile Reinforcement learning (aRL), in which the RL-agent performs informed exploration and executes only relevant actions. Consequently, the predictability of the RL-agent is enhanced, leading to rapid adaptation and convergence, which positions aRL as a suitable candidate for scheduling the tasks of soft real-time applications in edge computing. The experiments demonstrate that the combination of informed exploration and action-masking methods enables aRL to achieve a higher hit-ratio and converge faster than the baseline approaches.
\end{abstract}

\copyrightnotice

\section{Introduction}
\label{sec:intro}
In real-time systems, applications and tasks are classified as either `soft' or `hard' based on the consequences of deadline violations. In soft applications and tasks, deadline violations incur tolerable consequences, whereas in hard applications and tasks, such violations result in severe repercussions \cite{10197026}. Soft real-time applications (SRTAs) include multimedia streaming, online gaming, video surveillance, e-learning platforms, and social media feeds \cite{erickson2022soft}. Complex SRTAs impose significant processing demands in terms of computing, RAM, and storage resources. Consequently, executing these tasks on host devices, which are primarily resource-constrained embedded systems, while meeting strict timing requirements places a substantial burden on these devices. Hence, offloading portions of the SRTA workload to edge computing (EC) emerges as a viable solution \cite{goudarzi2022scheduling}.

Since portions of the workload of SRTAs are offloaded to the EC, the performance of SRTAs depends on the timeliness and accuracy of the task schedules generated by the edge scheduler. In addition, task scheduling in EC is NP-hard due to the need to optimize multiple objectives for both edge users and edge servers, evaluate numerous parameters inherent in EC components, and manage the dynamic EC environment \cite{wang2025tf}. Therefore, task scheduling methods based on heuristic and meta-heuristic algorithms face challenges in generating optimal or near-optimal task schedules owing to their inherent computational complexity, time-intensive calculations, requirement for fine-tuning numerous parameters, dependency on static EC models, and limited adaptability to dynamic EC conditions \cite{luo2021resource}. However, reinforcement learning (RL) effectively manages multiple parameters through high-dimensional state representations and adjusts its policies to optimize sequential decisions in dynamic environments. Subsequently, RL-based algorithms have demonstrated significant promise in addressing task scheduling challenges in EC \cite{hortelano2023comprehensive}.

RL algorithms offer considerable advantages for task scheduling in EC; however, prolonged learning times restrict their applicability to task scheduling for SRTAs in EC. This study investigates the potential impact of `informed exploration' and `action masking' on the learning time of an RL-agent. This study aims to minimize the time required by an RL-agent to generate task schedules while maximizing the accuracy of the generated schedules to achieve a high hit-ratio for SRTAs in EC. The distinct contributions of this study are summarized as follows: 1) A real-time-oriented formulation of the task scheduling problem in edge computing is introduced for soft real-time applications by incorporating a characterization of edge computing components based on resource availabilities and performance requirements. 2) A novel RL algorithm denoted aRL is proposed that (i) incorporates continuous action masking on the set of actions and states to identify relevant actions during the learning process, and (ii) employs an informed exploration strategy as an alternative to conventional exploration. 3) Experiments are conducted to evaluate the performance of aRL in various aspects, and baseline methods are employed for comparative analysis. Experimental results\footnote{The source code for the entire experiment and detailed results are available at \url{https://github.com/AminAvan/ARL-RT-TS-EC}.} demonstrate that aRL outperforms the baseline methods across multiple performance metrics under a consistent experimental setup and scenario.

The paper is structured as follows. Section \ref{sec:rltdwork} surveys recent RL-based edge schedulers in EC, and discusses the associated research challenges for scheduling SRTAs in EC. The system model and task scheduling mechanism of the proposed method are described in detail in Section \ref{sec:sysmdl}. The problem formulation for the proposed method is presented in Section \ref{sec:prblmfrml}, followed by Section \ref{sec:prpsdmthd}, which explains the aRL approach for addressing the task scheduling problem of SRTAs in EC environments. Section \ref{sec:exprmt} discusses the experimental results, and Section \ref{sec:cnclsn} concludes the paper.

\section{Related Work}
\label{sec:rltdwork}
EC has been instrumental in addressing the computational and memory demands of state-of-the-art SRTAs, which have become increasingly complex and resource-intensive while still possessing inherent timing constraints. The seamless execution of SRTAs in EC environments relies on effective task scheduling techniques that allocate offloaded tasks from edge users to appropriate edge servers. These techniques assign tasks based on the specific requirements of offloaded tasks and available resources of edge servers \cite{goudarzi2022scheduling,avan2023state}. EC comprises edge users and edge servers, where the latter include roadside units and base stations equipped with single-board computers, system-on-chip devices, server machines, access points, routers, and switches. The idle portions of the computational and memory resources in these devices are utilized within EC. As a distributed heterogeneous computing paradigm, EC presents significant challenges in task scheduling due to the numerous parameters associated with both edge users and edge servers that must be considered by task scheduling algorithms \cite{luo2021resource}. RL-based task scheduling leverages capability of RL to manage multiple parameters through high-dimensional state representations, dynamically refine policies, and optimize sequential decision-making in dynamic environments \cite{8771176}.

In \cite{he2024age}, the authors incorporated the Age of Information (AoI) to prioritize fresh environmental data and formulated the problem as a constrained Markov Decision Process (MDP) rather than a conventional MDP. They combined Post-Decision States (PDS) with a Deep Deterministic Policy Gradient (DDPG) to develop a novel task scheduling algorithm, termed Deep PDS (DPDS). Although DPDS improves the power consumption and AoI, it neglects the hit-ratio of offloaded tasks, which is a critical performance metric for SRTAs \cite{erickson2022soft}. Moreover, DPDS assumes the presence of a single edge server and optimizes the energy consumption of tasks based solely on the required processor cycles and data size. Furthermore, the experimental results indicated prolonged learning time for DPDS.

The DOSA algorithm in \cite{fan2023decentralized} addresses task scheduling by considering the communication costs and operational status of edge servers to reduce the latency. However, its reliance on RL results in a prolonged training time, thereby diminishing its suitability for SRTAs. Additionally, while DOSA accounts for processor cycles and data transmission sizes, it overlooks the heterogeneity of edge servers and neglects other critical factors such as remaining processing capacity, available RAM and storage volumes after task allocation, and power consumption criteria \cite{wang2025tf}. This oversight may lead to load imbalances and a reduced hit-ratio. Moreover, task performance is evaluated solely based on execution time, excluding critical metrics such as response time.

DCOM, a Deep RL (DRL) approach for task scheduling, was introduced to meet the Quality-of-Service (QoS) requirements of edge users \cite{geng2023deep}. This method aims to reduce both communication latency and power consumption in EC environments while explicitly accounting for the mobility of edge servers. However, DCOM encounters scalability challenges as the task volume increases. The resulting expansion of the state-action space for the RL-agent, combined with the complexity of modeling task dependencies through directed acyclic graphs (DAGs), contributes to a significantly prolonged learning time \cite{liu2024ga}.

A task scheduling technique was presented in \cite{liu2023asynchronous} to enhance overall EC performance by incorporating communication latency and bandwidth. This approach employs a DRL algorithm to optimize resource management and task scheduling across multiple edge servers, resulting in an optimal task schedule. In \cite{liu2022deep}, a DRL algorithm was proposed to reduce the execution time of computation-intensive tasks. This approach utilizes Deep Q-Networks (DQN) while integrating the computation and storage parameters of edge servers into the scheduling process. Although the authors considered multiple resource constraints, including computation, communication, and RAM, critical factors such as deadlines and response times were omitted. However, these parameters are essential for ensuring the effective performance of SRTAs \cite{erickson2022soft}.

In \cite{bansal2022urbanenqosplace}, a task scheduling approach based on the Dueling DQN was introduced for IoT applications in EC with the aim of minimizing both task execution time and power consumption for edge users. An online task scheduling technique employing the actor-critic framework of RL was introduced to address both stability and computational latency, with the goal of reducing overall power consumption throughout the EC infrastructure \cite{hoang2023deep}. A DDPG-based task scheduling technique that incorporates edge user mobility is proposed to reduce both the average execution time and power consumption in EC systems \cite{zhao2023meson}. In \cite{hsieh2023deep}, Double-DQN, Policy Gradient (PG), and actor-critic methods were developed and compared to optimize task execution time and reduce the task overflow rate. The actor-critic-based method outperformed the other approaches in achieving these objectives. In \cite{8657791}, a DQN-based task scheduling technique for EC was introduced to optimize both the average task execution time and the load balancing of edge servers. In \cite{8771176}, a DQN-based approach called DROO was introduced for task scheduling in EC to minimize a combined metric that reflects both the power consumption of the EC infrastructure and the task execution time of edge users. DROO employs a binary task scheduling procedure to address combinatorial optimization challenges inherent in task scheduling.

Most existing task scheduling techniques are designed for general-purpose applications, whereas SRTAs require additional metrics such as the hit-ratio, deadline, and response time \cite{8771176,he2024age,fan2023decentralized,wang2025tf,geng2023deep,liu2024ga,liu2023asynchronous,liu2022deep,bansal2022urbanenqosplace,hoang2023deep,zhao2023meson,hsieh2023deep,8657791}. These factors are as critical for SRTAs as power consumption, communication latency, execution time, and load balancing are for general-purpose scenarios. Existing literature frequently overlooks the runtime performance of scheduling algorithms \cite{8771176,he2024age,fan2023decentralized,wang2025tf,geng2023deep,liu2024ga,liu2023asynchronous,liu2022deep,bansal2022urbanenqosplace,hoang2023deep,zhao2023meson,hsieh2023deep,8657791}. Evaluations tend to disregard the time required for a scheduling algorithm to generate a schedule upon receiving input data or when changes occur in the EC environment. Specifically, in an RL-based approach, the runtime of the task scheduling algorithm is the learning time required for the RL-agent to converge, which differs from the inference time. The primary challenge associated with RL-based task scheduling for SRTAs is the prolonged learning time of RL algorithms. This prolonged learning time results from the agent’s reliance on trial-and-error exploration within an expansive state-action space \cite{he2024age,avan2023task}.

\section{System Model and Problem Formulation}
\label{sec:sysmdl}
\begin{figure}[h]
\centering
\includegraphics[width=\linewidth]{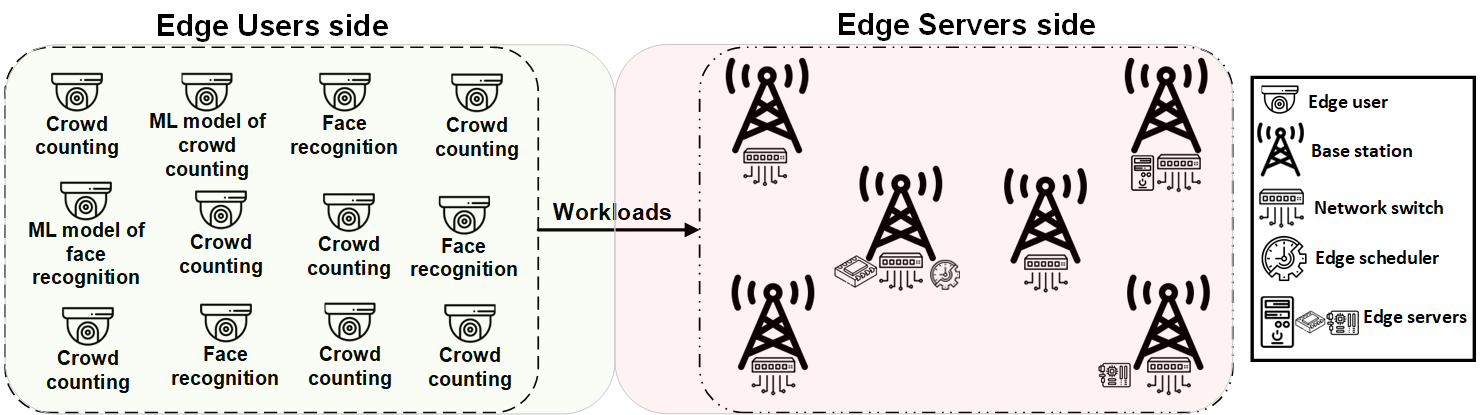}
\caption{Overview of the edge computing (EC) system model.}
\label{fig:sysmdl}
\end{figure}

This section presents an EC system supporting a video surveillance application, classified as an SRTA. As illustrated in Fig. \ref{fig:sysmdl}, each camera functions as an edge user and is responsible for one of the four services: 1) crowd counting, 2) face recognition, 3) machine learning (ML) model development for crowd counting, or 4) ML model development for face recognition. In addition, each edge server is one of the following computing platforms: Raspberry Pi, Jetson Nano, Jetson TX2, or an Intel Xeon processor (E5430, E5507, or E5645), as depicted in Fig. \ref{fig:sysmdl}. The cameras transmit their service characteristics to an edge scheduler located on one of the edge servers. Subsequently, the edge scheduler employs a task scheduling algorithm to allocate offloaded tasks across the available edge servers, leveraging the analysis of task characteristics and continuous monitoring of server resource capacity. However, both edge users and edge servers utilize individual system-level task scheduling algorithms to schedule their respective tasks.

\begin{figure}[h]
\centering
\includegraphics[width=0.6\linewidth]{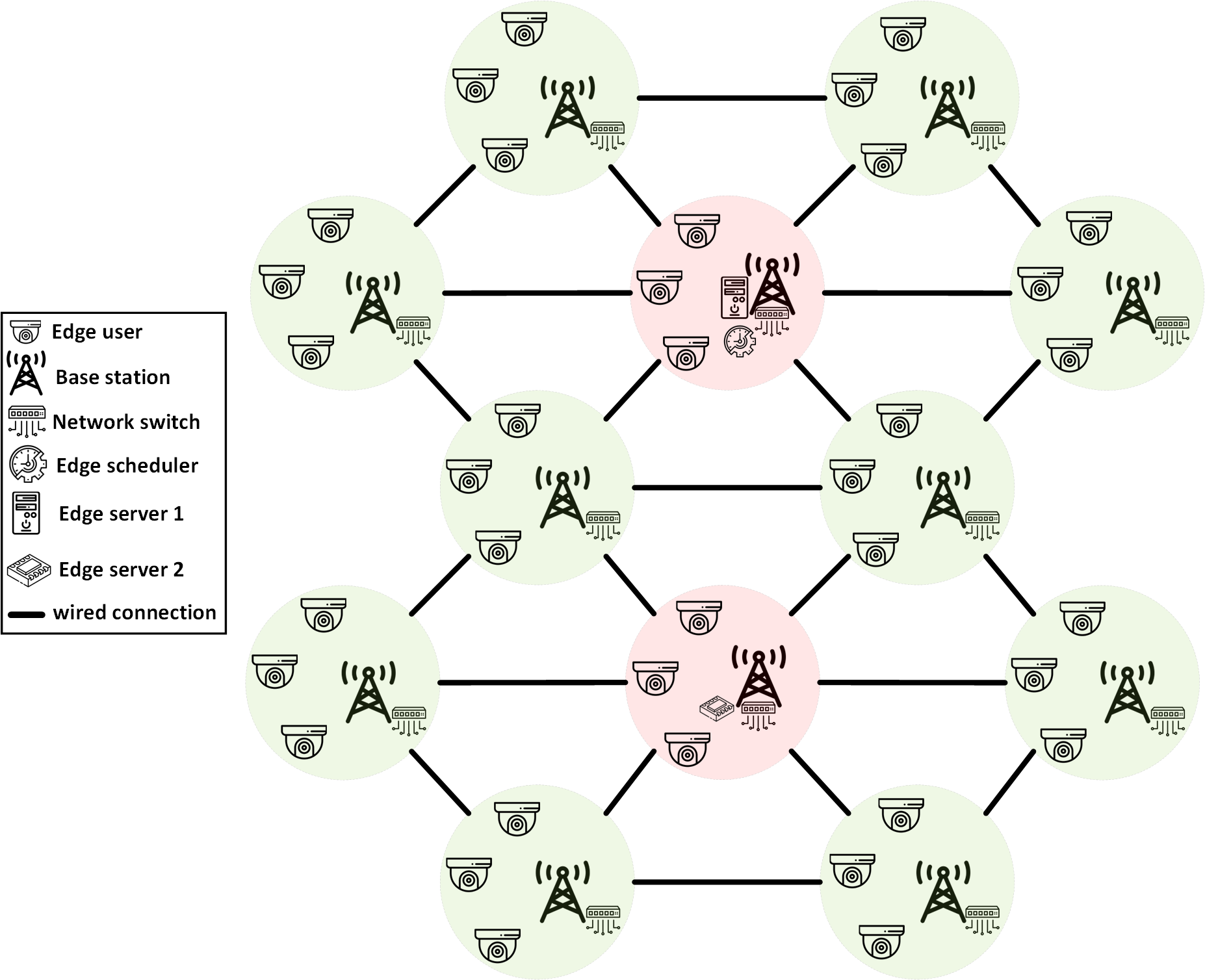}
\caption{Overview of the edge computing (EC) architecture.}
\label{fig:sysarch}
\end{figure}

In this study, an EC environment is composed of multiple base stations (BSs), each providing wireless connectivity to nearby edge users. These BSs are interconnected through wired links and network switches, forming a partial-mesh topology, as shown in Fig. \ref{fig:sysarch}. In Fig. \ref{fig:sysarch}, the zones highlighted in `green' represent BSs that serve solely as communication points, whereas zones highlighted in `red' indicate BSs with edge servers, enabling them to function as both communication points and computing nodes.

\textbf{Edge User Model:}
Edge users are denoted by $\mathcal{E} = \{\delta_1, \delta_2, \dots, \delta_n\}$, where each item $\delta_i$ is a camera, executing a specific service characterized by its workload $\Gamma_{\delta_i}$. The main goal of the video surveillance application in EC is considered counting the number of individuals in a monitored area, thereby rendering `crowd counting' is the most critical service of the application. In some instances where the identification of specific individuals is required, `face recognition' service is employed, thereby designating it as the second most critical service. In occasional instances, the development of ML models for both crowd counting and face recognition is required, which are considered the third and fourth most critical services, respectively. The development of ML models encompasses re-training, hyperparameter optimization, and fine-tuning. For each edge user, the service workload comprises tasks represented by $\tau_j \in \Gamma_{\delta_i}$. Each task is uniquely defined by a set of parameters, which include the arrival time ($a_j$), period ($p_j$), absolute deadline ($d_j$), required processor cycles ($c_j$), required RAM ($m_j$), and required storage ($l_j$). These parameters collectively characterize the task as follows:
\begin{equation}
\label{eq:tskchr}
    \tau_j=\{a_j, c_j, p_j, d_j, m_j, l_j\}
\end{equation}

\textbf{Edge Server Model:} Edge servers are defined by processor frequency $(\mathcal{F}_{s_k})$, number of processor cores $(\mathcal{N}_{s_k})$, RAM capacity $(\mathcal{M}_{s_k})$, and storage volume $(\mathcal{L}_{s_k})$. This representation enables task scheduling algorithms to support heterogeneous EC platforms. The set of edge servers is given by $\mathcal{S}=\{s_1, s_2, \ldots, s_k\}$, where each edge server $s_k \in \mathcal{S}$ is represented as:
\begin{equation}
\label{eq:srvchr}
s_k = \{\mathcal{F}_{s_k}, \mathcal{N}_{s_k}, \mathcal{M}_{s_k}, \mathcal{L}_{s_k}\}
\end{equation}

\subsection{Problem Formulation}
\label{sec:prblmfrml}
In deploying SRTAs within EC, the primary objective of the task scheduling algorithm employed by the edge scheduler is to generate a task schedule ($\Phi$), that maximizes the hit-ratio while minimizing the time required for its generation. The first constraint, which is essential for maximizing the hit-ratio of the task schedule ($\Phi$), requires that each task ($\tau_j \in \Gamma_{\delta_i}$) in the service workload of an edge user be allocated to an edge server ($s_k$) only if its response time ($r_{\tau_j:s_k}$) does not exceed its absolute deadline ($d_{\tau_j}$). Otherwise, the task will miss its deadline, and owing to sequential dependencies, subsequent tasks may also miss their deadlines.
\begin{equation}
\label{eq:frstcons}
C1.1:\quad\quad~r_{\tau_j:s_k} \leq d_{\tau_j}, \quad \forall \tau_j \in \Gamma_{\delta_i}
\end{equation}
The response time ($r_{\tau_j:s_k}$) of task ($\tau_j$) on edge server ($s_k$), is determined by several parameters, as expressed in Eq. \eqref{eq:dynmcexetme}. These parameters include the execution time on the hosting edge server ($e_{\tau_j}$), provisioning time for operational data ($t_{prov}$), round-trip time between the edge user and the hosting edge server ($t^{\delta_i:s_k}_{rtt}$), and inter-task communication time ($t^{\tau_j:\tau_x}_{rtt}$) arising in workloads with services distributed across multiple edge servers where task $\tau_j$ depends on the result of task $\tau_x$.
\begin{equation}
\label{eq:dynmcexetme}
r_{\tau_j:s_k} = e_{\tau_j}+t_{prov}+t^{\delta_i:s_k}_{rtt}+t^{\tau_j:\tau_x}_{rtt}
\end{equation}

The execution time of task ($e_{\tau_j}$) on edge server ($s_k$) is estimated by dividing the total required number of processor cycles by the processing capacity of the edge server.
\begin{equation}
\label{eq:exetme}
e_{\tau_j} = \frac{\text{total required processor cycles}}{\text{total processing capacity of edge server}} = \left(\frac{c_{\tau_j}\times m_{\tau_j}}{f_{s_k}\times N_{s_k}}\right)
\end{equation}

According to the second constraint defined in Eq. \eqref{eq:scndcons}, task allocation to an edge server is permitted if the edge server exhibits sufficient processing, RAM, and storage capacities.
\begin{equation}
\label{eq:scndcons}
C1.2:\quad\quad~((U_{P_{s_k}}+u_{p_{\tau_j}})<1)\land((U_{M_{s_k}}+u_{m_{\tau_j}})<1)\land((U_{\text{L}_{s_k}}+u_{l_{\tau_j}})<1)
\end{equation}

For each edge server $s_k$, processor utilization $U_{P_{s_k}}$ is computed as the sum of the individual task processor utilizations $u_{p_{\tau}}$ for all tasks hosted on the edge server. The processor utilization of each task ($u_{p_{\tau_j}}$) is determined by dividing its execution time ($e_{\tau_j}$) by its absolute deadline ($d_{\tau_j}$).
\begin{equation}
\label{eq:utlprcs}
U_{P_{s_k}} = \sum_{x=0}^{n}u_{p_{\tau_x}} = \sum_{x=0}^{n}\left(\frac{e_{\tau_x}}{d_{\tau_x}}\right)
\end{equation}

The hit-ratio of the generated task schedule ($\Phi$) is defined in Eq.\eqref{eq:funchitraio}, where $\mathcal{H}$ denotes the set of edge users for which every task in its workload satisfies its deadline. Accordingly, the objective of maximizing the hit-ratio is formulated as follows:
\begin{equation}
\label{eq:funchitraio}
f_{1}(\Phi) = \frac{|\mathcal{H}|}{|\mathcal{E}|}~,\quad(\mathcal{H} \subseteq \mathcal{E})
\end{equation}
\begin{equation}
\label{eq:objfnc1}
    \max\Bigl(f_{1}(\Phi)\Bigr)
\end{equation}
\hspace{2.7cm} $s.t.$
\begin{align}
\label{eq:objfnc1cns1}
C1.1: &\quad r_{\tau_j:s_k} \leq d_{\tau_j}, \quad \forall \tau_j \in \Gamma_{\delta_i} \\
\label{eq:objfnc1cns2}
C1.2: &\quad ((U_{P_{s_k}}+u_{p_{\tau_j}})<1)\land((U_{M_{s_k}}+u_{m_{\tau_j}})<1)\land((U_{\text{L}_{s_k}}+u_{l_{\tau_j}})<1)
\end{align}

To formulate the objective of minimizing the learning time of the RL-agent in generating task schedules for SRTAs, the first constraint requires that each task be exclusively assigned to a single edge server, as indicated in Eq. \eqref{eq:cns21}. However, the tasks of a workload are permitted to be assigned to different edge servers only if the assignment complies with constraints $C1.1$ and $C1.2$, as expressed in Eq. \eqref{eq:objfnc1cns1} and Eq. \eqref{eq:objfnc1cns2}, respectively.
\begin{equation}
\label{eq:cns21}
\sum_{k=1}^{\lvert\mathcal{S}\rvert} x_{\tau_js_k} \leq 1, \quad \forall\tau_j\in\delta_i,~\forall\delta_i\in\mathcal{E},~\forall s_k\in\mathcal{S}
\end{equation}

In addition, the second constraint for minimizing the learning time of RL-agent requires that the edge server $s_k$ be available only if the utilization of each resource remains below a predetermined threshold, as follows:
\begin{equation}
\label{eq:cns22}
(U_{P_{s_k}}\!\!< U_{\textnormal{th}}, U_{M_{s_k}}\!\!< U_{\textnormal{th}}, U_{L_{s_k}}\!\!< U_{\textnormal{th}}), \forall\, s_k \in \mathcal{S}
\end{equation}
The time required by the edge scheduler to identify an appropriate edge server for task $\tau_j$ and assign the task to that edge server is denoted by $t_\tau$. Individual times ($t_{\tau_1}, \ldots, t_{\tau_n}$) aggregate to form the overall runtime ($f_{2}(\Phi)$), as defined in Eq. \eqref{eq:fmultmsch}. Consequently, the objective is to minimize the runtime, as expressed in Eq. \eqref{eq:objfnc2}.
\begin{equation}
\label{eq:fmultmsch}
    f_{2}(\Phi) = \sum_{i=1}^{|\mathcal{E}|} \sum_{j=1}^{|\delta_i|} t_{\tau_j}, \quad \forall\tau_j\in\delta_i,~~~\forall\delta_i\in\mathcal{E}
\end{equation}
\begin{equation}
\label{eq:objfnc2}
    min\Bigl(f_{2}(\Theta)\Bigr)
\end{equation}
\hspace{0.5cm} $s.t.$
\begin{align}
\label{eq:objfnc2cns1}
C2.1: &\quad \sum_{k=1}^{\lvert\mathcal{S}\rvert} x_{\tau_js_k} \leq 1, \quad \forall\tau_j\!\in\!\delta_i, \forall\delta_i\!\in\!\mathcal{E}, \forall s_k\!\in\!\mathcal{S}\\
\label{eq:objfnc2cns2}
C2.2: &\quad (U_{P_{s_k}}\!\!< U_{\textnormal{th}}, U_{M_{s_k}}\!\!< U_{\textnormal{th}}, U_{L_{s_k}}\!\!< U_{\textnormal{th}}), \forall\, s_k \in \mathcal{S}
\end{align}

\section{aRL: Agile Reinforcement Learning}
\label{sec:prpsdmthd}
The primary objective is to minimize the runtime of the task scheduling algorithm when generating the task schedule while maximizing the hit-ratio of the resulting task schedule. Consequently, the task scheduling problem is formulated as an MDP, in which an action is represented as a two-element list, with the first element denoting the task and the second element representing the edge server, as follows:
\begin{equation}
a = [\tau_j, s_k]
\end{equation}
An RL-agent receives positive rewards to satisfy each of the four aforementioned constraints, including Eqs. \eqref{eq:objfnc1cns1},\eqref{eq:objfnc1cns2},\eqref{eq:objfnc2cns1},\eqref{eq:objfnc2cns2}, and negative rewards for each violation. Accordingly, the objective of the RL-agent is to identify a policy $(\pi)$ that maximizes cumulative rewards. Algorithm \ref{alg:ammdgp} details the process by which aRL generates task schedules and optimizes objectives, including the hit-ratio and learning time. As described in Section~\ref{sec:rltdwork}, the RL-agent experiences prolonged training times due to the expansive state-action space and inherent randomness during exploration. Consequently, Agile RL (aRL) employs informed exploration and action masking to accelerate learning and reduce learning time.

\begin{algorithm}
\footnotesize
\caption{Agile Reinforcement Learning (aRL)}
\label{alg:ammdgp}

\begin{algorithmic}[1]
\REQUIRE Set of $p$ offloaded tasks from edge users $\mathcal{T}=\{\tau_1, \tau_2, ..., \tau_p\}$ where ($\tau_j\in\Gamma_{\delta_i},~~\forall\delta_i\in\mathcal{E}$) \\
         \quad\quad Set of $q$ edge servers $\mathcal{S}=\{s_1, s_2, ..., s_q\}$, \quad $NumActionsTaken = 0$\\
\ENSURE Generated task schedule ($\Phi$)

\STATE \texttt{function GetEDFtask():}
\STATE \quad\quad $\Gamma_{ua} \leftarrow \operatorname{Sort}(\Gamma_{ua}) \text{ by } d_{\tau_j} \text{ in ascending order}$; return $\Gamma_{ua}[0]$

\STATE \texttt{function SelectAction(state):}
\STATE \quad\quad Compute $\varepsilon_{\text{threshold}}$;\enspace Compute $sample \leftarrow \operatorname{Random}(0,1)$;\enspace $\operatorname{Update}(\Gamma_{ua})$ based on Eq. \eqref{eq:snglasncns};
\STATE \quad\quad \textbf{if} $(sample>\varepsilon_{threshold})$ \textbf{then}
\STATE \quad\quad\quad $actionValues \gets \operatorname{policyNet}(state)$;\enspace $bestAction \gets \arg\max(actionValues)$;
\STATE \quad\quad\quad Return $\operatorname{mapAction}(bestAction)$
\STATE \quad\quad \textbf{else} Return $\operatorname{mapAction}(\texttt{GetEDFtask()})$

\STATE \textbf{begin main}

\FOR{$episode~\textbf{in}~range(episodes)$}
    \STATE $\mathbf{G} \gets \operatorname{matrix}(\mathbf{0}_{|\mathcal{T}| \times |\mathcal{S}|})$; \quad $t,reward,hitTasks\leftarrow0,0,0$; \quad $\Phi~\gets~\texttt{[]}$;
    \WHILE[{\tiny \quad\quad$\#\#$ `action bound' (Eq.\eqref{eq:lmtactns})}] {$\left(t \leq |\mathcal{T}|\right)~\textbf{or}~\left(hitTasks==|\mathcal{T}|\right)$} 
        \STATE $\Gamma_{ua} = \{ \tau_j \in \mathcal{T} : \sum_{k=1}^{|\mathcal{S}|} G_{j,k} = 0 \}$;
        \STATE $a_t=\texttt{SelectAction(state)}$; \text{\tiny{\quad\quad$\#\#$ $a_t:(\tau_j, s_k)$}}
        \STATE $t++$;
        \STATE $\textbf{if}~\left[(a_t(\tau_j)\in\Gamma_{ua}) == \texttt{TRUE}\right]~\textbf{then}$ {\tiny \quad\quad$\#\#$ `single-assignment constraint' (Eq.\eqref{eq:snglasncns})}
            \STATE \quad$reward~+=~positiveReward$;
            \STATE \quad$\textbf{if}~(U_{P{s_k},M{s_k},L{s_k}}<U_{th})~\textbf{and}~(U_{P{s_k},M{s_k},L{s_k}}+u_{p_{\tau_{j}},m_{\tau_{j},l_{\tau_{j}}}}<1)~\textbf{then}$ {\tiny \quad$\#\#$ edge server capacity constraints (Eqs.\eqref{eq:objfnc2cns2},\eqref{eq:objfnc1cns2})}
                \STATE \quad\quad$reward~+=~positiveReward$;
                    \STATE \quad\quad$\textbf{if}~(r_{\tau_j:s_k} \leq d_{\tau_j})~\textbf{then}$ {\tiny \quad\quad$\#\#$ response time against deadline (Eq.\eqref{eq:objfnc1cns1})}
                        \STATE \quad\quad\quad$reward~+=~positiveReward$;\enspace $hitTasks~+=~1$;
                    \STATE \quad\quad$\textbf{else}~reward~+=~negativeReward$;
            \STATE \quad$\textbf{else}~reward~+=~negativeReward$;
        \STATE $\textbf{else}~reward~+=~negativeReward$;
        \STATE $\Phi[t] \gets a_t$; \text{\tiny{\quad$\#\#$ append (task $\tau_j$, edge server $s_k$) pair to the task schedule}}
    \ENDWHILE
\ENDFOR
\STATE \textbf{end main}
\end{algorithmic}
\end{algorithm}

\textbf{aRL - Action Masking:} A decision matrix ($\mathbf{G}$), defined in Eq. \eqref{eq:exmmtrx}, tracks the assignment decisions of the RL-agent during each episode of the learning phase. At the beginning of each episode, all cells in the decision matrix ($\mathbf{G}$) are initialized to `zero'. When the RL-agent decides to assign a specific task to a particular edge server, the corresponding cell is updated to `one'. In decision matrix ($\mathbf{G}$), columns represent edge servers ($s_k \in \mathcal{S}$) and rows denote tasks ($\tau_j \in \Gamma_{\delta_i}$, where $\Gamma_{\delta_i} \subset \mathcal{E}$).

\begin{equation}
\label{eq:exmmtrx}
\mathbf{G}=
\begin{array}{c@{}c}
     & \begin{matrix} s_1 & \dots & s_m \end{matrix} \\
    \begin{matrix} \tau_1 \\ \vdots \\ \tau_n \end{matrix} &
    \left[\begin{array}{ccc}
    x_{11} & \dots & x_{1m} \\
    \vdots & \vdots & \vdots \\
    x_{n1} & \dots & x_{nm} \\
    \end{array}\right]
\end{array}
\end{equation}
\quad \quad \quad \quad \quad \quad $where:$
\begin{equation}
\label{eq:xvalue}
x_{\tau_js_k} =
\begin{cases}
1 & \text{RL-agent decides to assign task $\tau_j$ to edge server $s_k$} \\
0 & \text{otherwise}
\end{cases}
\end{equation}

As one mechanism to mask the actions of RL-agent and facilitate the execution of relevant actions, an `action bound' inspired by a finite-horizon concept is introduced. Under the action bound, the number of actions per episode is limited by the total number of tasks, as expressed in Eq. \eqref{eq:lmtactns}. This restriction mitigates excessively long rollouts and ensures that each episode terminates within a specified timeframe, thereby enabling more reliable policy and value function updates. Therefore, an episode terminates either when the total number of actions equals the total number of tasks or when all tasks are assigned to edge servers while satisfying the constraints in \eqref{eq:objfnc1cns1}, \eqref{eq:objfnc1cns2}, \eqref{eq:objfnc2cns1}, and \eqref{eq:objfnc2cns2}.
\begin{equation}
\label{eq:lmtactns}
|A|\leq|\mathcal{T}|
\end{equation}
\quad \quad \quad \quad \quad \quad $where:$
\begin{equation}
\label{eq:alltasks}
\mathcal{T}=\{\tau_1, \tau_2, ..., \tau_p\}    
\end{equation}
\begin{equation}
\label{eq:tactimpl}
\left(\sum_{j=1}^{n}\sum_{k=1}^{m}G_{j,k}\equiv|A|\right) ,\quad~A=\{a_1, a_2,\dots,a_{|\mathcal{T}|}\}
\end{equation}

The second mechanism, referred to as the `single-assignment constraint', restricts the RL-agent to one assignment decision per task, as defined in Eq. \eqref{eq:snglasncns}. After each action in an episode, the total number of possible actions (i.e., unassigned tasks) decreases, thereby reducing the state-action space over the course of the episode. In addition, the single-assignment constraint prevents the RL-agent from reassigning tasks that have already been assigned and directs its focus to the set of unassigned tasks ($\Gamma_{ua}$).
\begin{equation}
\label{eq:snglasncns}
\left(a = [\tau_j, s_k] \quad \forall \tau_j \in \Gamma_{ua} \right), \quad \Gamma_{ua} = \left\{ \tau_j \in \Gamma \subset \mathcal{E} \,\middle|\, \sum_{k=1}^{|\mathcal{S}|} G_{j,k} = 0 \right\}
\end{equation}

\textbf{aRL - Informed Exploration:} In exploration mode, aRL explores the action-state space using a modified version of the Earliest Deadline First (EDF) algorithm, adapted for EC, rather than relying solely on random exploration. As a result, aRL selects a task with the earliest deadline among unassigned tasks ($\Gamma_{ua}$) to be assigned to an available edge server subject to the conditions specified in Eqs. \eqref{eq:objfnc1cns2} and \eqref{eq:objfnc2cns2}.

\textbf{aRL - Reward Function:} The reward function in aRL provides positive rewards for actions that satisfy the constraints in \eqref{eq:objfnc1cns1}, \eqref{eq:objfnc1cns2}, \eqref{eq:objfnc2cns1}, and \eqref{eq:objfnc2cns2}. Otherwise, each violation of any constraint incurs a negative reward. When the RL-agent takes the action $a_t = (\tau_j, s_k)$, which assigns task $\tau_j$ to edge server $s_k$, multiple conditions based on the aforementioned constraints are evaluated to determine the corresponding rewards. According to constraint \eqref{eq:objfnc1cns2}, a positive reward is granted if task $\tau_j$ is an unassigned task (i.e., $\tau_j \in \Gamma_{ua}$). In line with constraints \eqref{eq:objfnc2cns2} and \eqref{eq:objfnc1cns2}, another positive reward is awarded if $s_k$ has a sufficient capacity. The criticality of the task is then evaluated, and an additional positive reward is assigned accordingly. In this study, the criticality levels of tasks in crowd counting, face recognition, ML model development for crowd counting, and ML model development for face recognition are designated as highest, second, third, and fourth, respectively. Finally, in accordance with constraint \eqref{eq:objfnc1cns1}, the response time $(r_{\tau_j:s_k})$ is assessed; if it meets the deadline $d_{\tau_j}$ (i.e., $r_{\tau_j:s_k} \leq d_{\tau_j}$), a positive reward is granted. However, if the response time exceeds the task deadline (i.e., $r_{\tau_j:s_k} > d_{\tau_j}$), the RL-agent receives a negative reward because the task misses its deadline ($d_{\tau_j}$) with this action $a_t = (\tau_j, s_k)$.

\section{Experimental Results and Analysis}
\label{sec:exprmt}
In this paper, a video surveillance application in EC is evaluated, wherein resource-constrained cameras serve as edge users executing dedicated services and edge servers with diverse resource characteristics are available. The system comprises $22$ zones, each served by a base station (BS) and a network switch. BSs are connected via wired links through network switches, forming a partially meshed topology. Two BSs are equipped with NVIDIA Jetson TX2 devices, while the other two use an Intel Xeon E5430 and an Intel Xeon E5645 processor, respectively. A total of $52$ edge users are randomly distributed, each wirelessly connected to the base station in its zone; $44$ perform crowd counting, six execute face recognition, and one is assigned to ML-model development for crowd counting while another is assigned to ML-model development for face recognition. Each service comprises specific tasks, each with distinct requirements regarding processor cycles per megabyte of data, available RAM, and storage capacity. The aRL and baseline methods are implemented in EdgeSimPy, an open-source EC simulator that provides detailed abstractions for EC components \cite{souza2023edgesimpy}. The methods are evaluated at `runtime' in terms of hit-ratio, RAM usage, and power consumption. The `runtime' is defined as the total time required to generate an optimal or near-optimal schedule for tasks offloaded by SRTAs in EC, or to adapt an existing task schedule to environmental changes, such as variations in the number of edge users, their tasks, or edge servers, as well as the time required to provision tasks to the edge servers according to that task schedule. Runtime is measured for the task scheduling algorithm because the generated schedule targets edge users with SRTAs that have specific timing requirements. Additionally, RAM and power consumption are evaluated since EC environments have limited resources compared to cloud computing, making resource efficiency essential. The runtime for the heuristic task scheduling algorithm is defined as the total time required to generate a task schedule and provision tasks accordingly. In contrast, the runtime for the RL-based task scheduling algorithm is defined as the total time required to complete its learning phase and achieve convergence based on inputs provided by EC components, rather than merely the inference time. The RL-based task scheduling algorithms, including the aRL and Vanilla RL (vRL) approaches, are implemented using a DQN architecture with a fully connected feedforward neural network and utilize a three-layer structure to approximate the Q-value function for task scheduling decisions in EC. vRL employs the same framework and architecture as aRL but omits the informed exploration and action masking mechanisms integrated into aRL. Each of the RL-based task scheduling algorithms is executed $31$ times independently to validate the results\footnote{The detailed logs and plots for $31$ runs are available in the `\texttt{detailed\_results}' directory at \url{https://github.com/AminAvan/ARL-RT-TS-EC/tree/main/detailed_results}.}, and the average hit-ratio, runtime, RAM usage, and power consumption are computed \cite{avan2023task}. Convergence is considered to be achieved if the hit-ratio of the generated task schedule exceeds $98\%$ for $100$ consecutive episodes and remains stable above this threshold \cite{yu2019convergent}.

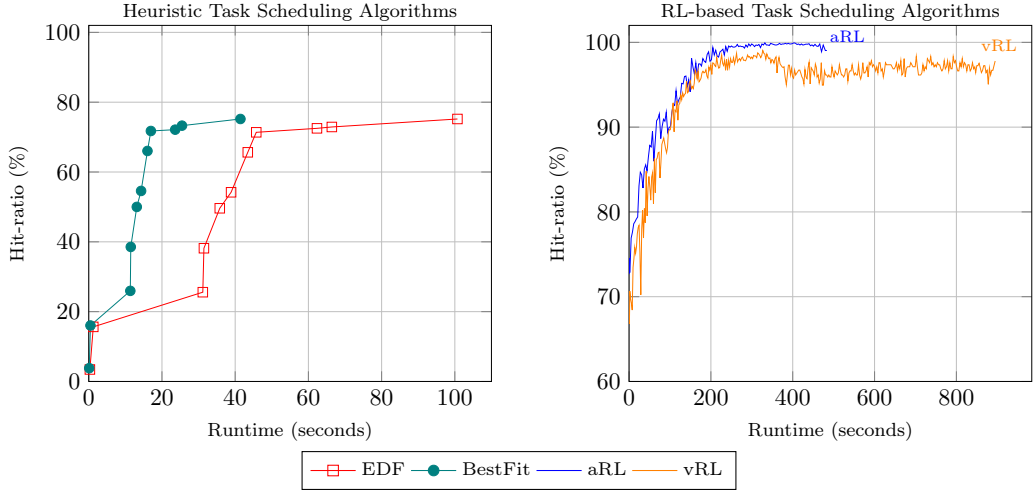
\begin{figure}[h]
    \centering
    \begin{minipage}[b]{0.49\linewidth}
        \centering
        \begin{tikzpicture}[scale=0.9]
    \begin{axis}[
        title={Heuristic Task Scheduling Algorithms},
        title style={font=\footnotesize},
        title style={yshift=-7.5pt},
        width=1.1\linewidth,
        height=\linewidth,
        xlabel={{\footnotesize Runtime (seconds)}},
        ylabel={{\footnotesize Hit-ratio (\%)}},
        xmin=0, xmax=110,
        ymin=0, ymax=102,
        legend style={at={(0.5,-0.15)},
        anchor=north,legend columns=-1},
        legend to name=named,
        grid=major,
        ymajorgrids=true,
        xmajorgrids=true,
    ]
    \addplot[color=red,mark=square] coordinates {
        (0.31, 3.43) (1.243, 15.64) (31.156, 25.57) (31.46, 38.16) (35.79, 49.61) (38.88, 54.19) (43.44, 65.64) (45.76, 71.37) (62.34, 72.51) (66.4, 72.90) (100.713, 75.19)
    };

    \addplot[color=teal,mark=*] coordinates {
        (0.13, 3.81) (0.47, 16.03) (11.353, 25.95) (11.48, 38.54) (13.123, 50) (14.3, 54.58) (16.04, 66.03) (16.97, 71.75) (23.61, 72.13) (25.47, 73.28) (41.435, 75.19)
    };
    
    \addplot[color=blue] coordinates {
      (0,0)
    };

    \addplot[color=orange] coordinates {
    (0,0)
    };

    \legend{EDF, BestFit, aRL, vRL}
    \node[anchor=west] at (rel axis cs:1, 0.5) {Text on Right Edge};
    \end{axis}
\end{tikzpicture}
    \end{minipage}
    \hfill
    \begin{minipage}[b]{0.49\linewidth}
        \centering
        \begin{tikzpicture}[scale=0.9]
    \begin{axis}
    [
        title={RL-based Task Scheduling Algorithms},
        title style={font=\footnotesize},
        title style={yshift=-7.5pt},
        width=1.1\linewidth,
        height=\linewidth,
        xlabel={{\footnotesize Runtime (seconds)}},
        ylabel={{\footnotesize Hit-ratio (\%)}},
        xmin=0,
        ymin=60,
        ymax=102,
        legend style={at={(0.5,-0.15)},
        anchor=north,legend columns=-1},
        grid=major,
        ymajorgrids=true,
        xmajorgrids=true,
    ]

    \addplot[color=blue] coordinates {
(0.48, 73.33)
(0.91, 74.5)
(1.34, 72.83)
(2.03, 72.89)
(5.13, 76.92)
(8.27, 77.61)
(11.43, 78.6)
(14.62, 78.85)
(17.82, 79.16)
(21.05, 79.4)
(24.28, 83.07)
(27.52, 84.68)
(30.78, 84.37)
(34.05, 82.82)
(37.33, 85.05)
(40.62, 85.55)
(43.94, 84.86)
(47.26, 86.48)
(50.6, 87.84)
(53.93, 87.66)
(57.28, 89.52)
(60.63, 85.98)
(63.99, 88.34)
(67.34, 90.76)
(70.7, 91.0)
(74.06, 91.56)
(77.44, 88.65)
(80.81, 90.07)
(84.19, 90.94)
(87.58, 90.7)
(90.95, 91.75)
(94.36, 89.39)
(97.77, 89.95)
(101.19, 90.07)
(104.61, 91.75)
(108.04, 92.68)
(111.46, 92.56)
(114.89, 94.29)
(118.3, 92.74)
(121.74, 93.24)
(125.19, 93.55)
(128.63, 95.16)
(132.06, 95.1)
(135.49, 95.97)
(138.95, 96.03)
(142.4, 94.79)
(145.85, 95.6)
(149.29, 94.17)
(152.74, 98.14)
(156.19, 97.08)
(159.65, 96.28)
(163.13, 97.58)
(166.6, 96.03)
(170.07, 97.46)
(173.54, 97.46)
(177.0, 97.21)
(180.45, 98.08)
(183.92, 97.77)
(187.39, 97.64)
(190.89, 98.57)
(194.35, 98.88)
(197.83, 97.77)
(201.29, 97.95)
(204.8, 99.38)
(208.31, 98.51)
(211.8, 99.07)
(215.31, 98.39)
(218.8, 98.26)
(222.3, 98.88)
(225.8, 99.01)
(229.31, 99.19)
(232.83, 98.51)
(236.33, 99.5)
(239.85, 98.95)
(243.37, 99.57)
(246.86, 99.44)
(250.37, 99.57)
(253.88, 99.44)
(257.39, 99.44)
(260.92, 99.5)
(264.42, 99.75)
(267.96, 99.5)
(271.48, 99.57)
(275.02, 99.5)
(278.54, 99.5)
(282.04, 99.32)
(285.57, 99.69)
(289.09, 99.44)
(292.61, 99.81)
(296.15, 99.57)
(299.71, 99.81)
(303.25, 99.69)
(306.78, 99.5)
(310.31, 99.69)
(313.91, 99.63)
(317.43, 99.75)
(321.22, 99.38)
(324.78, 99.81)
(328.33, 99.69)
(331.92, 99.94)
(335.5, 99.69)
(339.08, 99.69)
(342.68, 99.63)
(346.25, 99.88)
(349.84, 99.81)
(353.44, 99.75)
(356.99, 99.69)
(360.53, 99.69)
(364.11, 99.75)
(367.67, 99.88)
(371.24, 99.75)
(374.81, 99.81)
(378.37, 99.81)
(381.96, 99.88)
(385.53, 99.81)
(389.1, 99.88)
(392.7, 99.81)
(396.29, 99.81)
(399.85, 99.88)
(403.44, 99.94)
(407.02, 99.88)
(410.58, 99.75)
(414.19, 99.81)
(417.78, 99.69)
(421.35, 99.81)
(425.02, 99.68)
(428.67, 99.73)
(432.28, 99.79)
(436.07, 99.78)
(439.81, 99.84)
(443.4, 99.6)
(447.03, 99.7)
(450.66, 99.55)
(454.42, 99.76)
(458.36, 99.56)
(462.07, 99.52)
(466.05, 99.81)
(470.44, 98.9)
(474.57, 99.68)
(479.71, 99.04)
(483.39, 99.04)
    }
    node[pos=1.1, above] {{\scriptsize aRL}};
    
    \addplot[color=orange] coordinates {
    (0.46, 67.12)
(0.86, 66.81)
(1.27, 68.3)
(1.84, 70.66)
(3.79, 69.73)
(5.75, 69.23)
(7.74, 68.42)
(9.76, 73.45)
(11.8, 74.44)
(13.85, 75.81)
(15.93, 75.12)
(18.05, 75.81)
(20.16, 77.42)
(22.28, 78.16)
(24.4, 78.41)
(26.55, 75.19)
(28.73, 70.22)
(30.9, 77.98)
(33.09, 80.21)
(35.28, 76.92)
(37.49, 80.4)
(39.69, 78.72)
(41.9, 84.8)
(44.1, 79.53)
(46.32, 82.82)
(48.53, 84.18)
(50.76, 82.94)
(52.98, 81.39)
(55.21, 84.12)
(57.44, 84.61)
(59.68, 80.96)
(61.93, 86.29)
(64.19, 82.26)
(66.45, 86.17)
(68.71, 86.35)
(71.01, 86.97)
(73.29, 87.04)
(75.57, 84.0)
(77.84, 85.79)
(80.13, 87.96)
(82.41, 88.34)
(84.7, 88.71)
(86.99, 88.09)
(89.32, 87.47)
(91.61, 86.97)
(93.91, 87.59)
(96.2, 89.45)
(98.51, 89.7)
(100.83, 89.7)
(103.15, 90.45)
(105.45, 92.0)
(107.75, 92.87)
(110.07, 89.45)
(112.38, 92.0)
(114.69, 91.81)
(117.01, 93.11)
(119.33, 90.82)
(121.65, 92.62)
(123.97, 92.8)
(126.3, 93.8)
(128.64, 93.24)
(130.98, 94.04)
(133.3, 93.61)
(135.66, 93.86)
(138.01, 93.98)
(140.36, 95.41)
(142.72, 94.05)
(145.07, 94.17)
(147.45, 95.22)
(149.81, 95.47)
(152.16, 94.54)
(154.54, 94.79)
(156.89, 94.54)
(159.27, 95.16)
(161.64, 96.71)
(164.01, 96.34)
(166.39, 95.72)
(168.76, 96.59)
(171.15, 95.35)
(173.54, 95.53)
(175.91, 95.78)
(178.3, 96.4)
(180.68, 96.59)
(183.07, 96.4)
(185.47, 96.46)
(187.85, 96.77)
(190.25, 96.03)
(192.63, 95.78)
(195.03, 97.08)
(197.42, 96.71)
(199.81, 97.33)
(202.22, 96.22)
(204.61, 97.58)
(207.02, 97.52)
(209.42, 97.27)
(211.81, 97.15)
(214.23, 98.2)
(216.63, 97.46)
(219.04, 96.96)
(221.45, 97.02)
(223.86, 97.83)
(226.27, 98.51)
(228.7, 97.52)
(231.1, 97.27)
(233.52, 97.89)
(235.94, 96.71)
(238.35, 98.02)
(240.78, 96.71)
(243.2, 97.89)
(245.63, 97.08)
(248.07, 98.08)
(250.5, 97.64)
(252.93, 97.64)
(255.37, 98.2)
(257.79, 97.46)
(260.23, 97.95)
(262.67, 98.7)
(265.11, 97.7)
(267.54, 97.95)
(270.0, 98.08)
(272.42, 97.83)
(274.87, 98.14)
(277.32, 98.26)
(279.75, 97.89)
(282.24, 97.89)
(284.71, 97.83)
(287.17, 98.08)
(289.66, 98.51)
(292.13, 98.26)
(294.59, 98.32)
(297.08, 98.08)
(299.56, 98.26)
(302.02, 98.08)
(304.5, 98.45)
(306.99, 98.39)
(309.46, 98.39)
(311.95, 98.2)
(314.45, 98.76)
(316.93, 98.63)
(319.42, 98.45)
(321.91, 98.2)
(324.35, 98.46)
(326.84, 99.04)
(329.35, 98.72)
(331.76, 98.61)
(334.24, 98.01)
(336.64, 98.64)
(339.14, 98.32)
(341.57, 98.08)
(344.09, 98.16)
(346.62, 97.59)
(349.12, 97.98)
(351.65, 97.88)
(354.19, 97.54)
(356.69, 98.08)
(359.2, 98.29)
(361.63, 96.39)
(364.07, 97.69)
(366.48, 98.22)
(369.01, 97.48)
(371.55, 97.28)
(374.05, 97.11)
(376.42, 97.03)
(378.72, 97.12)
(381.21, 95.96)
(384.43, 95.0)
(386.94, 96.15)
(389.45, 97.31)
(391.96, 97.31)
(394.49, 96.16)
(397.0, 96.92)
(399.53, 95.96)
(402.08, 95.77)
(404.6, 96.73)
(407.12, 96.15)
(409.65, 96.35)
(412.18, 96.54)
(414.7, 96.73)
(417.24, 96.73)
(419.81, 96.35)
(422.28, 95.94)
(424.79, 95.72)
(427.33, 95.08)
(429.88, 96.8)
(432.41, 95.94)
(434.95, 95.72)
(437.54, 97.01)
(440.06, 97.65)
(442.6, 95.09)
(445.15, 97.01)
(447.7, 96.79)
(450.23, 96.58)
(452.78, 96.15)
(455.38, 97.44)
(457.93, 97.01)
(459.55, 95.91)
(462.11, 96.15)
(464.71, 95.19)
(467.27, 95.67)
(469.83, 97.6)
(472.4, 94.95)
(474.97, 94.95)
(477.75, 96.64)
(480.46, 96.15)
(483.02, 96.15)
(485.6, 95.91)
(488.17, 96.15)
(490.75, 95.91)
(493.37, 95.91)
(495.93, 97.12)
(498.51, 96.39)
(501.1, 95.91)
(503.7, 96.63)
(506.26, 96.39)
(508.85, 96.16)
(511.46, 96.39)
(514.06, 97.36)
(516.65, 96.64)
(519.25, 96.16)
(521.83, 96.15)
(524.43, 95.91)
(527.0, 96.15)
(529.59, 96.87)
(532.22, 96.88)
(534.81, 95.43)
(537.41, 96.39)
(540.01, 96.64)
(542.61, 96.63)
(545.2, 97.36)
(547.78, 96.64)
(550.37, 96.15)
(553.02, 96.39)
(555.61, 96.15)
(558.22, 96.16)
(560.83, 97.36)
(563.44, 96.15)
(566.05, 95.67)
(568.64, 96.16)
(571.26, 97.6)
(573.91, 96.39)
(576.52, 96.39)
(579.15, 96.64)
(581.77, 97.12)
(584.38, 95.91)
(587.02, 97.6)
(589.62, 97.36)
(592.25, 97.84)
(594.9, 97.12)
(597.53, 97.36)
(600.16, 97.11)
(602.82, 96.63)
(605.44, 97.12)
(608.14, 96.39)
(610.77, 97.6)
(613.43, 97.6)
(616.11, 96.63)
(618.77, 96.87)
(621.44, 95.91)
(624.07, 96.63)
(626.74, 97.6)
(629.41, 97.35)
(632.07, 97.36)
(634.72, 97.36)
(637.38, 97.36)
(640.08, 97.36)
(642.74, 97.12)
(645.43, 96.87)
(648.1, 96.88)
(650.76, 97.12)
(653.46, 97.6)
(656.12, 96.63)
(658.83, 97.36)
(661.53, 96.88)
(664.22, 97.84)
(666.9, 96.15)
(669.59, 96.39)
(672.28, 98.08)
(674.95, 95.91)
(677.66, 96.88)
(680.34, 97.12)
(683.04, 97.12)
(685.76, 97.12)
(688.47, 97.6)
(691.17, 97.36)
(693.86, 97.12)
(696.56, 97.12)
(699.27, 97.12)
(702.0, 98.32)
(704.69, 97.11)
(707.4, 96.15)
(710.14, 96.15)
(712.85, 96.15)
(715.54, 96.39)
(718.26, 97.36)
(720.99, 96.64)
(723.67, 97.12)
(726.41, 96.88)
(729.12, 98.08)
(731.84, 97.84)
(734.59, 97.6)
(737.31, 98.32)
(740.01, 97.36)
(742.75, 97.6)
(745.49, 97.12)
(748.2, 97.36)
(750.95, 97.12)
(753.67, 97.6)
(756.38, 97.36)
(759.12, 97.35)
(761.9, 96.88)
(764.63, 96.63)
(767.38, 96.88)
(770.25, 97.84)
(772.97, 96.87)
(775.73, 97.12)
(778.48, 97.84)
(781.21, 96.63)
(783.97, 97.6)
(786.75, 96.88)
(789.51, 96.88)
(792.24, 98.32)
(795.0, 97.6)
(797.77, 97.84)
(800.51, 97.12)
(803.28, 96.63)
(806.04, 96.63)
(808.78, 97.6)
(811.56, 97.12)
(814.36, 98.08)
(817.13, 97.12)
(819.91, 97.6)
(822.66, 97.36)
(825.45, 96.63)
(828.21, 97.36)
(831.01, 97.36)
(833.78, 96.63)
(836.54, 96.88)
(839.33, 97.6)
(842.14, 96.88)
(844.93, 97.11)
(847.12, 96.43)
(849.89, 97.25)
(852.66, 97.25)
(855.45, 97.53)
(858.4, 96.98)
(861.17, 96.43)
(863.94, 96.7)
(866.75, 97.25)
(869.58, 97.25)
(872.38, 96.43)
(875.16, 96.7)
(877.96, 95.05)
(880.77, 96.98)
(883.55, 96.15)
(886.33, 96.7)
(889.14, 96.7)
(891.93, 97.25)
(894.74, 97.8)
    }
    node[pos=1.01] {{\scriptsize vRL}};
    \end{axis}
\end{tikzpicture}

    \end{minipage}
    \vspace{0.2em}
    \ref{named}
\caption{Hit-ratio versus runtime analysis.}
\label{fig:rntmanlys}
\end{figure}

Figure \ref{fig:rntmanlys} illustrates that aRL achieves task scheduling hit-ratios exceeding those of EDF, BestFit, and vRL by $24\%$, $24\%$, and $1.9\%$, respectively, with aRL being $46\%$ faster than vRL. Consequently, aRL is capable of scheduling the offloaded tasks of SRTAs in EC.

\begin{figure}[h]
    \centering
    \begin{minipage}[b]{0.49\linewidth}
        \centering
        \begin{tikzpicture}[scale=0.9]
    \begin{axis}[
        title={Heuristic Task Scheduling Algorithms},
        title style={font=\footnotesize},
        title style={yshift=-7.5pt},
        width=1.1\linewidth,
        height=\linewidth,
        xlabel={{\footnotesize RAM usage (MB)}},
        ylabel={{\footnotesize Hit-ratio (\%)}},
        xmin=0,
        ymin=0, ymax=102,
        legend style={at={(0.5,-0.15)},
        anchor=north,legend columns=-1},
        legend to name=named,
        grid=major,
        ymajorgrids=true,
        xmajorgrids=true,
    ]
    \addplot[color=red,mark=square] coordinates {
        (82.07, 3.43) (331.23, 15.64) (9895.06, 25.57) (10025.06, 38.16) (11765.42, 49.61) (13035.97, 54.19) (14962.05, 65.64) (16044.93, 71.37) (49867.98, 75.19)
    };

    \addplot[color=teal,mark=*] coordinates {
        (81.74, 3.81) (329.73, 16.03) (9875.37, 25.95) (10005.11, 38.54) (11745.07, 50) (13013.96, 54.58) (14938.38, 66.03) (16022.02, 71.75) (24345.76, 72.9) (49834.18, 75.19)
    };
    

    \addplot[color=blue] coordinates {
      (0,0)
    };

    \addplot[color=orange] coordinates {
    (0,0)
    };
    
    \legend{EDF, BestFit, aRL, vRL}
    \end{axis}
\end{tikzpicture}
    \end{minipage}
    \hfill
    \begin{minipage}[b]{0.49\linewidth}
        \centering
        \input{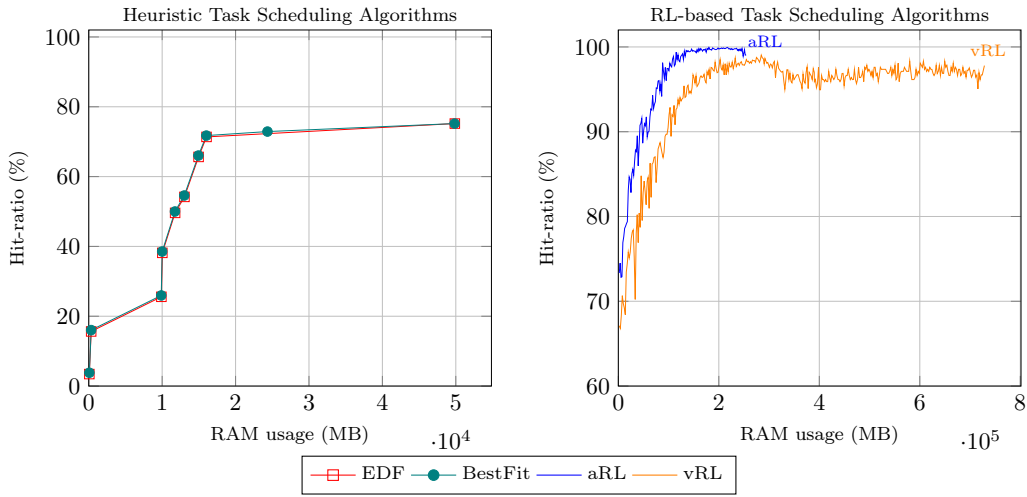}
    \end{minipage}
    \vspace{0.2em}
    \ref{named}
\caption{Hit-ratio versus RAM usage analysis.}
\label{fig:hrmemusg}
\end{figure}

Figure \ref{fig:hrmemusg} demonstrates that aRL achieves the hit-ratio objective while utilizing $65.5\%$ less RAM than vRL, whereas EDF and BestFit are unable to achieve the hit-ratio objective. Accordingly, aRL, which exhibits reduced RAM usage compared to vRL and the highest hit-ratio among all evaluated methods, imposes limited RAM-related overhead in EC while satisfying the timing requirements of edge users.

\begin{figure}[h]
\centering
    \begin{minipage}[b]{0.49\linewidth}
        \centering
        \begin{tikzpicture}[scale=0.9]
    \begin{axis}[
        title={Heuristic Task Scheduling Algorithms},
        title style={font=\footnotesize},
        title style={yshift=-7.5pt},
        width=1.1\linewidth,
        height=\linewidth,
        xlabel={{\tiny Power consumption (Watt-seconds)}},
        ylabel={{\footnotesize Hit-ratio (\%)}},
        xmin=0,
        ymin=0, ymax=102,
        legend style={at={(0.5,-0.15)},
        anchor=north,legend columns=-1},
        legend to name=named,
        grid=major,
        ymajorgrids=true,
        xmajorgrids=true,
    ]
    \addplot[color=red,mark=square] coordinates {
        (21.37, 3.43) (61.48, 15.64) (1101.08, 25.57) (1109.14, 38.16) (1215.72, 49.61) (1306.57, 54.19) (1419.09, 65.64) (1715.51, 71.37) (3747.89, 75.19)
    };

    \addplot[color=teal,mark=*] coordinates {
        (5, 3.81) (32.46, 16.03) (941.3, 25.95) (958.39, 38.54) (1089.98, 50) (1172.65, 54.58) (1568.54, 66.03) (1787.59, 71.75) (3382.04, 72.9) (5942.51, 75.19)
    };
    

    \addplot[color=blue] coordinates {
    (0,0)
    };

    \addplot[color=orange] coordinates {
    (0,0)
    };
    
    \legend{EDF, BestFit, aRL, vRL}
    \end{axis}
\end{tikzpicture}
    \end{minipage}
    \hfill
    \begin{minipage}[b]{0.49\linewidth}
        \centering
        \begin{tikzpicture}[scale=0.9]
    \begin{axis}[
        title={RL-based Task Scheduling Algorithms},
        title style={font=\footnotesize},
        title style={yshift=-7.5pt},
        width=1.1\linewidth,
        height=\linewidth,
        xlabel={{\tiny Power consumption (Watt-seconds)}},
        ylabel={{\footnotesize Hit-ratio (\%)}},
        xmin=0,
        ymin=60,
        ymax=102,
        legend style={at={(0.5,-0.15)},
        anchor=north,legend columns=-1},
        grid=major,
        ymajorgrids=true,
        xmajorgrids=true,
    ]

    \addplot[color=blue] coordinates {
    (33.17, 73.33)
(69.27, 74.5)
(107.63, 72.83)
(172.91, 72.89)
(236.97, 76.92)
(297.73, 77.61)
(362.42, 78.6)
(426.3, 78.85)
(490.72, 79.16)
(554.5, 79.4)
(616.44, 83.07)
(677.83, 84.68)
(740.26, 84.37)
(801.84, 82.82)
(864.77, 85.05)
(927.7, 85.55)
(986.7, 84.86)
(1045.11, 86.48)
(1104.7, 87.84)
(1160.49, 87.66)
(1222.61, 89.52)
(1281.74, 85.98)
(1342.95, 88.34)
(1406.74, 90.76)
(1465.43, 91.0)
(1523.79, 91.56)
(1582.66, 88.65)
(1646.08, 90.07)
(1704.23, 90.94)
(1763.86, 90.7)
(1822.0, 91.75)
(1880.32, 89.39)
(1940.05, 89.95)
(1997.83, 90.07)
(2060.17, 91.75)
(2118.85, 92.68)
(2178.75, 92.56)
(2237.85, 94.29)
(2300.32, 92.74)
(2359.28, 93.24)
(2418.68, 93.55)
(2478.0, 95.16)
(2533.07, 95.1)
(2593.34, 95.97)
(2651.75, 96.03)
(2708.85, 94.79)
(2770.15, 95.6)
(2830.78, 94.17)
(2887.84, 98.14)
(2945.66, 97.08)
(3006.56, 96.28)
(3065.47, 97.58)
(3124.7, 96.03)
(3186.27, 97.46)
(3245.81, 97.46)
(3305.45, 97.21)
(3367.65, 98.08)
(3424.94, 97.77)
(3485.11, 97.64)
(3544.7, 98.57)
(3606.27, 98.88)
(3664.01, 97.77)
(3724.27, 97.95)
(3784.04, 99.38)
(3842.77, 98.51)
(3901.95, 99.07)
(3960.63, 98.39)
(4020.13, 98.26)
(4080.03, 98.88)
(4139.3, 99.01)
(4194.69, 99.19)
(4252.61, 98.51)
(4313.95, 99.5)
(4372.14, 98.95)
(4431.23, 99.57)
(4489.78, 99.44)
(4548.6, 99.57)
(4605.43, 99.44)
(4660.28, 99.44)
(4720.14, 99.5)
(4778.73, 99.75)
(4834.57, 99.5)
(4892.62, 99.57)
(4951.57, 99.5)
(5009.26, 99.5)
(5066.68, 99.32)
(5124.42, 99.69)
(5180.08, 99.44)
(5240.03, 99.81)
(5298.39, 99.57)
(5354.59, 99.81)
(5409.08, 99.69)
(5467.54, 99.5)
(5525.0, 99.69)
(5584.54, 99.63)
(5646.48, 99.75)
(5706.52, 99.38)
(5764.34, 99.81)
(5818.96, 99.69)
(5877.42, 99.94)
(5935.06, 99.69)
(5990.27, 99.69)
(6047.19, 99.63)
(6106.42, 99.88)
(6162.98, 99.81)
(6217.74, 99.75)
(6272.9, 99.69)
(6329.86, 99.69)
(6386.02, 99.75)
(6443.98, 99.88)
(6501.76, 99.75)
(6556.88, 99.81)
(6613.17, 99.81)
(6669.37, 99.88)
(6726.38, 99.81)
(6786.01, 99.88)
(6845.24, 99.81)
(6903.38, 99.81)
(6960.44, 99.88)
(7018.94, 99.94)
(7076.72, 99.88)
(7134.55, 99.75)
(7191.25, 99.81)
(7246.45, 99.69)
(7306.22, 99.81)
(7359.15, 99.68)
(7422.1, 99.73)
(7480.1, 99.79)
(7533.52, 99.78)
(7591.02, 99.84)
(7650.16, 99.6)
(7705.47, 99.7)
(7770.66, 99.55)
(7822.88, 99.76)
(7869.91, 99.56)
(7924.83, 99.52)
(7972.68, 99.81)
(8015.6, 98.9)
(8071.73, 99.68)
(8129.45, 99.04)
(8183.9, 99.04)
    }
    node[pos=1.15, above] {{\scriptsize aRL}};
    
    \addplot[color=orange] coordinates {
(37.19, 67.12)
(78.71, 66.81)
(119.6, 68.3)
(191.06, 70.66)
(265.87, 69.73)
(331.24, 69.23)
(399.05, 68.42)
(468.79, 73.45)
(539.85, 74.44)
(610.73, 75.81)
(677.59, 75.12)
(749.01, 75.81)
(816.9, 77.42)
(886.74, 78.16)
(954.1, 78.41)
(1019.42, 75.19)
(1084.38, 70.22)
(1149.03, 77.98)
(1212.77, 80.21)
(1278.68, 76.92)
(1346.89, 80.4)
(1411.85, 78.72)
(1473.34, 84.8)
(1537.94, 79.53)
(1603.12, 82.82)
(1667.41, 84.18)
(1730.06, 82.94)
(1799.41, 81.39)
(1863.74, 84.12)
(1923.91, 84.61)
(1991.08, 80.96)
(2049.5, 86.29)
(2111.79, 82.26)
(2173.46, 86.17)
(2239.64, 86.35)
(2300.22, 86.97)
(2365.59, 87.04)
(2429.96, 84.0)
(2488.1, 85.79)
(2554.42, 87.96)
(2616.4, 88.34)
(2677.34, 88.71)
(2740.41, 88.09)
(2805.05, 87.47)
(2865.77, 86.97)
(2929.87, 87.59)
(2992.48, 89.45)
(3054.28, 89.7)
(3118.52, 89.7)
(3180.5, 90.45)
(3246.73, 92.0)
(3308.71, 92.87)
(3371.55, 89.45)
(3429.46, 92.0)
(3492.66, 91.81)
(3554.92, 93.11)
(3615.95, 90.82)
(3679.69, 92.62)
(3740.27, 92.8)
(3799.5, 93.8)
(3859.94, 93.24)
(3920.57, 94.04)
(3980.61, 93.61)
(4044.03, 93.86)
(4105.56, 93.98)
(4166.37, 95.41)
(4226.63, 94.05)
(4287.75, 94.17)
(4348.79, 95.22)
(4410.99, 95.47)
(4469.27, 94.54)
(4528.32, 94.79)
(4587.59, 94.54)
(4650.25, 95.16)
(4714.62, 96.71)
(4775.47, 96.34)
(4838.22, 95.72)
(4899.48, 96.59)
(4960.92, 95.35)
(5021.81, 95.53)
(5083.84, 95.78)
(5147.35, 96.4)
(5208.3, 96.59)
(5262.37, 96.4)
(5322.73, 96.46)
(5380.55, 96.77)
(5441.41, 96.03)
(5500.36, 95.78)
(5561.62, 97.08)
(5623.78, 96.71)
(5682.6, 97.33)
(5739.16, 96.22)
(5797.57, 97.58)
(5857.07, 97.52)
(5917.15, 97.27)
(5980.04, 97.15)
(6037.46, 98.2)
(6097.99, 97.46)
(6157.49, 96.96)
(6214.68, 97.02)
(6275.22, 97.83)
(6334.99, 98.51)
(6396.92, 97.52)
(6456.1, 97.27)
(6512.21, 97.89)
(6569.09, 96.71)
(6630.03, 98.02)
(6687.95, 96.71)
(6747.67, 97.89)
(6807.3, 97.08)
(6863.5, 98.08)
(6922.32, 97.64)
(6983.35, 97.64)
(7043.08, 98.2)
(7101.26, 97.46)
(7162.97, 97.95)
(7222.02, 98.7)
(7283.5, 97.7)
(7343.77, 97.95)
(7400.96, 98.08)
(7463.3, 97.83)
(7524.56, 98.14)
(7584.37, 98.26)
(7643.96, 97.89)
(7705.26, 97.89)
(7762.01, 97.83)
(7822.59, 98.08)
(7879.96, 98.51)
(7937.61, 98.26)
(7999.81, 98.32)
(8057.1, 98.08)
(8115.01, 98.26)
(8172.25, 98.08)
(8229.26, 98.45)
(8291.2, 98.39)
(8348.39, 98.39)
(8403.83, 98.2)
(8465.94, 98.76)
(8528.06, 98.63)
(8584.21, 98.45)
(8641.45, 98.2)
(8701.31, 98.46)
(8761.01, 99.04)
(8819.82, 98.72)
(8872.61, 98.61)
(8946.84, 98.01)
(9016.35, 98.64)
(9076.69, 98.32)
(9130.29, 98.08)
(9189.35, 98.16)
(9253.33, 97.59)
(9310.44, 97.98)
(9367.55, 97.88)
(9439.42, 97.54)
(9500.83, 98.08)
(9563.88, 98.29)
(9611.34, 96.39)
(9689.45, 97.69)
(9756.42, 98.22)
(9811.97, 97.48)
(9855.3, 97.28)
(9915.93, 97.11)
(9996.27, 97.03)
(10051.62, 97.12)
(10110.34, 95.96)
(10166.12, 95.0)
(10220.5, 96.15)
(10274.88, 97.31)
(10328.7, 97.31)
(10388.12, 96.16)
(10449.22, 96.92)
(10509.76, 95.96)
(10570.02, 95.77)
(10631.54, 96.73)
(10690.54, 96.15)
(10750.24, 96.35)
(10807.98, 96.54)
(10867.4, 96.73)
(10926.96, 96.73)
(10986.66, 96.35)
(11068.98, 95.94)
(11123.47, 95.72)
(11183.09, 95.08)
(11238.51, 96.8)
(11298.13, 95.94)
(11359.62, 95.72)
(11416.76, 97.01)
(11471.56, 97.65)
(11530.24, 95.09)
(11592.2, 97.01)
(11653.22, 96.79)
(11709.27, 96.58)
(11769.51, 96.15)
(11817.47, 97.44)
(11870.56, 97.01)
(11922.23, 95.91)
(11981.22, 96.15)
(12042.67, 95.19)
(12094.68, 95.67)
(12151.23, 97.6)
(12204.8, 94.95)
(12264.68, 94.95)
(12311.95, 96.64)
(12371.48, 96.15)
(12430.65, 96.15)
(12486.85, 95.91)
(12549.17, 96.15)
(12610.8, 95.91)
(12666.48, 95.91)
(12726.17, 97.12)
(12787.8, 96.39)
(12838.57, 95.91)
(12891.97, 96.63)
(12950.45, 96.39)
(13007.7, 96.16)
(13064.6, 96.39)
(13121.5, 97.36)
(13179.97, 96.64)
(13236.35, 96.16)
(13292.38, 96.15)
(13351.9, 95.91)
(13404.42, 96.15)
(13462.55, 96.87)
(13517.87, 96.88)
(13573.02, 95.43)
(13633.77, 96.39)
(13690.15, 96.64)
(13746.52, 96.63)
(13801.33, 97.36)
(13858.92, 96.64)
(13911.28, 96.15)
(13970.45, 96.39)
(14024.9, 96.15)
(14084.95, 96.16)
(14141.85, 97.36)
(14195.6, 96.15)
(14254.08, 95.67)
(14313.6, 96.16)
(14369.62, 97.6)
(14426.7, 96.39)
(14489.2, 96.39)
(14543.12, 96.64)
(14599.85, 97.12)
(14658.33, 95.91)
(14714.18, 97.6)
(14770.2, 97.36)
(14823.95, 97.84)
(14880.85, 97.12)
(14936.35, 97.36)
(14991.5, 97.11)
(15043.33, 96.63)
(15101.98, 97.12)
(15160.1, 96.39)
(15217.0, 97.6)
(15277.93, 97.6)
(15339.38, 96.63)
(15391.2, 96.87)
(15442.5, 95.91)
(15498.7, 96.63)
(15554.2, 97.6)
(15607.42, 97.35)
(15657.85, 97.36)
(15718.6, 97.36)
(15774.8, 97.36)
(15827.32, 97.36)
(15883.52, 97.12)
(15934.12, 96.87)
(15985.95, 96.88)
(16040.22, 97.12)
(16089.25, 97.6)
(16142.13, 96.63)
(16192.9, 97.36)
(16240.35, 96.88)
(16295.85, 97.84)
(16351.17, 96.15)
(16400.55, 96.39)
(16456.92, 98.08)
(16508.92, 95.91)
(16557.78, 96.88)
(16608.9, 97.12)
(16660.9, 97.12)
(16715.52, 97.12)
(16767.7, 97.6)
(16817.08, 97.36)
(16869.6, 97.12)
(16927.02, 97.12)
(16967.3, 97.12)
(17014.75, 98.32)
(17070.78, 97.11)
(17128.55, 96.15)
(17188.6, 96.15)
(17239.55, 96.15)
(17292.43, 96.39)
(17347.23, 97.36)
(17397.65, 96.64)
(17453.67, 97.12)
(17507.25, 96.88)
(17561.7, 98.08)
(17616.33, 97.84)
(17668.5, 97.6)
(17722.6, 98.32)
(17774.6, 97.36)
(17833.08, 97.6)
(17876.15, 97.12)
(17930.6, 97.36)
(17982.25, 97.12)
(18037.22, 97.6)
(18088.35, 97.36)
(18139.3, 97.35)
(18195.67, 96.88)
(18246.98, 96.63)
(18300.02, 96.88)
(18353.42, 97.84)
(18411.9, 96.87)
(18459.53, 97.12)
(18516.6, 97.84)
(18568.95, 96.63)
(18625.5, 97.6)
(18682.58, 96.88)
(18730.38, 96.88)
(18786.75, 98.32)
(18831.58, 97.6)
(18882.35, 97.84)
(18934.17, 97.12)
(18990.9, 96.63)
(19042.72, 96.63)
(19097.87, 97.6)
(19152.15, 97.12)
(19201.88, 98.08)
(19260.53, 97.12)
(19315.67, 97.6)
(19365.92, 97.36)
(19422.83, 96.63)
(19480.78, 97.36)
(19535.75, 97.36)
(19589.5, 96.63)
(19643.25, 96.88)
(19694.38, 97.6)
(19754.08, 96.88)
(19805.02, 97.11)
(19878.0, 96.43)
(19935.0, 97.25)
(19982.6, 97.25)
(20040.2, 97.53)
(20090.6, 96.98)
(20144.2, 96.43)
(20195.6, 96.7)
(20246.6, 97.25)
(20294.8, 97.25)
(20351.0, 96.43)
(20403.4, 96.7)
(20458.8, 95.05)
(20517.6, 96.98)
(20564.0, 96.15)
(20622.0, 96.7)
(20675.2, 96.7)
(20733.2, 97.25)
(20784.0, 97.8)
    }
    node[pos=1.01] {{\scriptsize vRL}};

    \end{axis}
\end{tikzpicture}
    \end{minipage}
    \vspace{0.2em}
    \ref{named}
\caption{Hit-ratio versus power consumption analysis.}
\label{fig:hrpwrcns}
\end{figure}
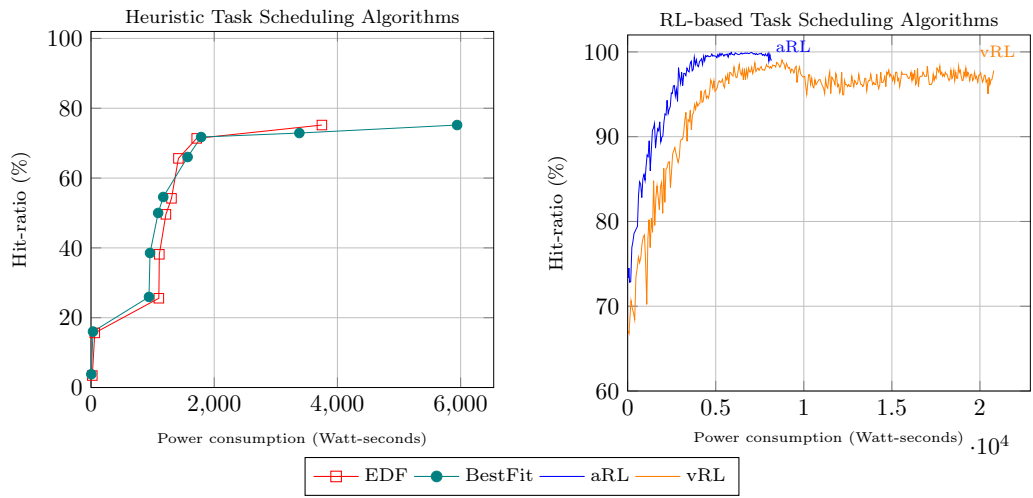

Figure \ref{fig:hrpwrcns} indicates that aRL accomplishes the hit-ratio objective while consuming $60.6\%$ less power than vRL, whereas EDF and BestFit fail to attain the hit-ratio objective. Therefore, aRL is a power-efficient, RL-based task scheduling algorithm for SRTAs in EC.

\section{Conclusion}
\label{sec:cnclsn}
Complex soft real-time applications can be executed by edge users with limited resources using edge computing. However, scheduling offloaded tasks across edge servers in EC is challenging due to heterogeneous edge server characteristics, the timing constraints of edge users, and the dynamic nature of edge computing. Consequently, heuristic and metaheuristic task scheduling algorithms struggle to produce optimal or near-optimal schedules because of limited adaptability to dynamic conditions and complex environments. Reinforcement learning approaches have emerged as promising candidates owing to their self-learning capabilities and online policy updates through environmental interaction. Nevertheless, when applied to medium- and large-scale problems, reinforcement learning approaches suffer from prolonged learning times. Therefore, a novel reinforcement learning approach, termed Agile Reinforcement Learning (aRL), is proposed to address the timing constraints of soft real-time applications and reduce the extended learning time associated with reinforcement learning. This approach incorporates `informed exploration' and `action masking' to prioritize relevant actions and accelerate convergence. Experimental results demonstrate that aRL attains the highest hit-ratio among the heuristic and reinforcement learning baseline methods, with reductions of $46\%$ in runtime, $65.5\%$ in RAM usage, and $60.6\%$ in power consumption relative to the vRL approach.

\printbibliography[heading=subbibintoc]

@article{goudarzi2022scheduling,
  title={Scheduling IoT applications in edge and fog computing environments: a taxonomy and future directions},
  author={Goudarzi, Mohammad and Palaniswami, Marimuthu and Buyya, Rajkumar},
  journal={ACM Computing Surveys},
  volume={55},
  number={7},
  pages={1--41},
  year={2022},
  publisher={ACM New York, NY}
}

@incollection{erickson2022soft,
  title={Soft real-time scheduling},
  author={Erickson, Jeremy P and Anderson, James H},
  booktitle={Handbook of Real-Time Computing},
  pages={233--267},
  year={2022},
  publisher={Springer}
}

@article{luo2021resource,
  title={Resource scheduling in edge computing: A survey},
  author={Luo, Quyuan and Hu, Shihong and Li, Changle and Li, Guanghui and Shi, Weisong},
  journal={IEEE Communications Surveys \& Tutorials},
  volume={23},
  number={4},
  pages={2131--2165},
  year={2021},
  publisher={IEEE}
}

@article{hortelano2023comprehensive,
  title={A comprehensive survey on reinforcement-learning-based computation offloading techniques in edge computing systems},
  author={Hortelano, Diego and de Miguel, Ignacio and Barroso, Ram{\'o}n J Dur{\'a}n and Aguado, Juan Carlos and Merayo, Noem{\'\i} and Ruiz, Lidia and Asensio, Adrian and Masip-Bruin, Xavi and Fern{\'a}ndez, Patricia and Lorenzo, Rub{\'e}n M and others},
  journal={Journal of Network and Computer Applications},
  volume={216},
  pages={103669},
  year={2023},
  publisher={Elsevier}
}

@article{avan2023state,
  title={A state-of-the-art review of task scheduling for edge computing: A delay-sensitive application perspective},
  author={Avan, Amin and Azim, Akramul and Mahmoud, Qusay H},
  journal={Electronics},
  volume={12},
  number={12},
  pages={2599},
  year={2023},
  publisher={MDPI}
}

@ARTICLE{8771176,
  author={Huang, Liang and Bi, Suzhi and Zhang, Ying-Jun Angela},
  journal={IEEE Transactions on Mobile Computing}, 
  title={Deep Reinforcement Learning for Online Computation Offloading in Wireless Powered Mobile-Edge Computing Networks}, 
  year={2020},
  volume={19},
  number={11},
  pages={2581-2593},
  doi={10.1109/TMC.2019.2928811}}

@article{he2024age,
  title={Age-Based Scheduling for Mobile Edge Computing: A Deep Reinforcement Learning Approach},
  author={He, Xingqiu and You, Chaoqun and Quek, Tony QS},
  journal={IEEE Transactions on Mobile Computing},
  year={2024},
  publisher={IEEE}
}

@article{wang2025tf,
  title={TF-DDRL: A Transformer-enhanced Distributed DRL Technique for Scheduling IoT Applications in Edge and Cloud Computing Environments},
  author={Wang, Zhiyu and Goudarzi, Mohammad and Buyya, Rajkumar},
  journal={IEEE Transactions on Services Computing},
  year={2025},
  publisher={IEEE}
}

@article{fan2023decentralized,
  title={Decentralized Scheduling for Concurrent Tasks in Mobile Edge Computing via Deep Reinforcement Learning},
  author={Fan, Ye and Ge, Jidong and Zhang, Sheng and Wu, Jie and Luo, Bin},
  journal={IEEE Transactions on Mobile Computing},
  year={2023},
  publisher={IEEE}
}

@article{geng2023deep,
  title={Deep Reinforcement Learning Based Distributed Computation Offloading in Vehicular Edge Computing Networks},
  author={Geng, Liwei and Zhao, Hongbo and Wang, Jiayue and Kaushik, Aryan and Yuan, Shuai and Feng, Wenquan},
  journal={IEEE Internet of Things Journal},
  year={2023},
  publisher={IEEE}
}

@article{liu2024ga,
  title={GA-DRL: Graph Neural Network-Augmented Deep Reinforcement Learning for DAG Task Scheduling over Dynamic Vehicular Clouds},
  author={Liu, Zhang and Huang, Lianfen and Gao, Zhibin and Luo, Manman and Hosseinalipour, Seyyedali and Dai, Huaiyu},
  journal={IEEE Transactions on Network and Service Management},
  year={2024},
  publisher={IEEE}
}

@article{liu2023asynchronous,
  title={Asynchronous Deep Reinforcement Learning for Collaborative Task Computing and On-Demand Resource Allocation in Vehicular Edge Computing},
  author={Liu, Lei and Feng, Jie and Mu, Xuanyu and Pei, Qingqi and Lan, Dapeng and Xiao, Ming},
  journal={IEEE Transactions on Intelligent Transportation Systems},
  year={2023},
  publisher={IEEE}
}

@article{liu2022deep,
  title={Deep reinforcement learning based approach for online service placement and computation resource allocation in edge computing},
  author={Liu, Tong and Ni, Shenggang and Li, Xiaoqiang and Zhu, Yanmin and Kong, Linghe and Yang, Yuanyuan},
  journal={IEEE Transactions on Mobile Computing},
  year={2022},
  publisher={IEEE}
}

@article{bansal2022urbanenqosplace,
  title={Urbanenqosplace: A deep reinforcement learning model for service placement of real-time smart city iot applications},
  author={Bansal, Maggi and Chana, Inderveer and Clarke, Siobh{\'a}n},
  journal={IEEE Transactions on Services Computing},
  volume={16},
  number={4},
  pages={3043--3060},
  year={2022},
  publisher={IEEE}
}

@article{hoang2023deep,
  title={Deep reinforcement learning-based online resource management for uav-assisted edge computing with dual connectivity},
  author={Hoang, Linh T and Nguyen, Chuyen T and Pham, Anh T},
  journal={IEEE/ACM Transactions on Networking},
  volume={31},
  number={6},
  pages={2761--2776},
  year={2023},
  publisher={IEEE}
}

@article{zhao2023meson,
  title={MESON: A mobility-aware dependent task offloading scheme for urban vehicular edge computing},
  author={Zhao, Liang and Zhang, Enchao and Wan, Shaohua and Hawbani, Ammar and Al-Dubai, Ahmed Y and Min, Geyong and Zomaya, Albert Y},
  journal={IEEE Transactions on Mobile Computing},
  volume={23},
  number={5},
  pages={4259--4272},
  year={2023},
  publisher={IEEE}
}

@article{hsieh2023deep,
  title={Deep reinforcement learning-based task assignment for cooperative mobile edge computing},
  author={Hsieh, Li-Tse and Liu, Hang and Guo, Yang and Gazda, Robert},
  journal={IEEE Transactions on Mobile Computing},
  volume={23},
  number={4},
  pages={3156--3171},
  year={2023},
  publisher={IEEE}
}

@ARTICLE{8657791,
  author={Wang, Jiadai and Zhao, Lei and Liu, Jiajia and Kato, Nei},
  journal={IEEE Transactions on Emerging Topics in Computing}, 
  title={Smart Resource Allocation for Mobile Edge Computing: A Deep Reinforcement Learning Approach}, 
  year={2021},
  volume={9},
  number={3},
  pages={1529-1541},
  doi={10.1109/TETC.2019.2902661}}

@inproceedings{avan2023task,
  title={A Task Scheduler for Mobile Edge Computing Using Priority-based Reinforcement Learning},
  author={Avan, Amin and Kheiri, Farnaz and Mahmoud, Qusay H and Azim, Akramul and Makrehchi, Masoud and Rahnamayan, Shahryar},
  booktitle={2023 IEEE Symposium Series on Computational Intelligence (SSCI)},
  pages={539--546},
  year={2023},
  organization={IEEE}
}

@article{souza2023edgesimpy,
    author={Paulo S. Souza and Tiago Ferreto and Rodrigo N. Calheiros},
    title={EdgeSimPy: Python-Based Modeling and Simulation of Edge Computing Resource Management Policies},
    journal={Future Generation Computer Systems},
    year={2023},
    issn={0167-739X},
    volume={148},
    pages={446-459},
    doi={https://doi.org/10.1016/j.future.2023.06.013},
    publisher={Elsevier}
}

@article{yu2019convergent,
  title={Convergent policy optimization for safe reinforcement learning},
  author={Yu, Ming and Yang, Zhuoran and Kolar, Mladen and Wang, Zhaoran},
  journal={Advances in Neural Information Processing Systems},
  volume={32},
  year={2019}
}

@INPROCEEDINGS{10197026,
  author={Avan, Amin and Azim, Akramul and Mahmoud, Qusay H.},
  booktitle={2023 IEEE 26th International Symposium on Real-Time Distributed Computing (ISORC)}, 
  title={A Robust Scheduling Algorithm for Overload-Tolerant Real-Time Systems}, 
  year={2023},
  volume={},
  number={},
  pages={1-10},
  doi={10.1109/ISORC58943.2023.00013}}

\end{document}